\documentclass{article}

\usepackage{PRIMEarxiv}

\usepackage{natbib}
\usepackage{amsmath}
\usepackage{amssymb}
\usepackage{algorithm}
\usepackage{algorithmic}

\usepackage[utf8]{inputenc} % allow utf-8 input
\usepackage[T1]{fontenc}    % use 8-bit T1 fonts
\usepackage{hyperref}       % hyperlinks
\usepackage{url}            % simple URL typesetting
\usepackage{booktabs}       % professional-quality tables
\usepackage{amsfonts}       % blackboard math symbols
\usepackage{nicefrac}       % compact symbols for 1/2, etc.
\usepackage{microtype}      % microtypography
\usepackage{lipsum}
\usepackage{fancyhdr}       % header
\usepackage{graphicx}       % graphics
\graphicspath{{media/}}     % organize your images and other figures under media/ folder

\title{Parameter Efficient Continual Learning for Sparse Event-Based Transformers}

\author{
  Vaishnavi Nagabhushana\\
  SustainAI Lab, MFSDS\&AI\\
  IIT Guwahati\\
  \texttt{n.vaishnavi@iitg.ac.in} \\
    \And
  Kartikay Agrawal\\
  SustainAI Lab, MFSDS\&AI\\
  IIT Guwahati\\
  \texttt{a.kartikay@iitg.ac.in} \\
     \And
  Ayon Borthakur\\
  SustainAI Lab, MFSDS\&AI\\
  IIT Guwahati\\
  \texttt{ayon.borthakur@iitg.ac.in}
}
\begin{document}
\maketitle

\begin{abstract}
Robotic and edge intelligence systems operate in dynamic environments where data arrives continuously, requiring models to adapt while preserving previously learned knowledge under strict memory and energy constraints. While parameter-efficient fine-tuning has shown promise for continual learning with vision transformers, conventional architectures rely on dense computation and remain costly for real-world deployment. Sparse event-based vision transformers provide energy-efficient event-driven computation, yet their continual learning capabilities remain largely unexplored. We here introduce \textbf{sLoTh}, a parameter-efficient continual learning framework for pretrained sparse event-based (spiking) vision transformers. sLoTh freezes the backbone and restricts plasticity to scalable-efficient low-rank attention updates (\textbf{seLoRA}) and shared neuronal \textbf{threshold modulation}, enabling adaptation without replay buffers by updating \textbf{less than 1\%} of model parameters. Experiments across CIFAR-100, Tiny-ImageNet, ImageNet-100, and ImageNet-R with up to \textbf{100 tasks} demonstrate competitive rehearsal-free performance in class-incremental learning and online continual learning, while enabling approximately \textbf{6.5$\times$} lower energy consumption than conventional dense vision transformers.
\end{abstract}

% keywords can be removed
% \keywords{First keyword \and Second keyword \and More}

% \section{TLDR}
% The paper introduces a sLoTh, a parameter-efficient replay-free Continual Learning framework for Spiking Transformers. The method is efficient, updating fewer than 1\% of the total parameters. 

\section{Introduction}
\begin{figure}[ht]
    \centering
    \includegraphics[width=0.9\textwidth]{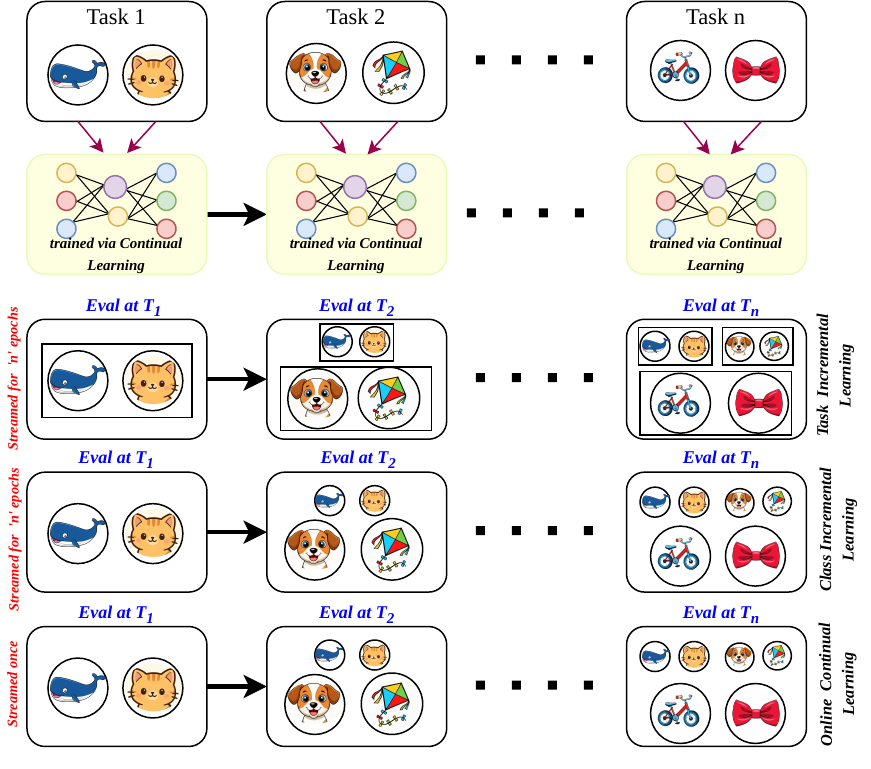}
    \caption{Comparison of Continual Learning Paradigms. In Task-Incremental Learning (\textbf{TIL}), \textbf{task boundaries are known}, allowing for task-specific components during evaluation. Class-Incremental Learning (\textbf{CIL}) increases difficulty by requiring the model to distinguish between all seen classes \textbf{without task IDs} at inference. Online Continual Learning (\textbf{OCL}) is a \textbf{challenging scenario}, as it combines the expanding output space of CIL with the constraint that data is streamed only once.}
    \label{fig:three paradigms of cl}
\end{figure}

Deep neural networks achieve remarkable performance on large static datasets but struggle when data arrives incrementally. In such settings described in figure \ref{fig:three paradigms of cl}, models suffer from \emph{\textbf{catastrophic forgetting}} \cite{threeparadiagms, ewc}, where learning new tasks overwrites previously acquired knowledge due to unconstrained plasticity. Continual Learning (CL) addresses this challenge by enabling models to learn sequentially while retaining prior knowledge \cite{AGEM,ewc,si}. Recent years have seen a strong shift from CNNs to vision transformers \cite{vit} for downstream tasks due to their transferable pretrained representations. However, naively fine-tuning pretrained transformers in continual learning leads to severe forgetting and high computational overhead \cite{sdlora, ptm-survey}. Parameter-Efficient Fine-Tuning (PEFT) methods, including prompt tuning \cite{l2p,wang2022dualprompt}, adapters \cite{adaptercil}, and low-rank adaptation (LoRA) \cite{hu2022lora}, mitigate this issue by updating only a small subset of parameters while freezing the pretrained backbone. These approaches substantially reduce memory and computation requirements while maintaining strong continual learning performance \cite{zhang2023adaLoRA,liang2025gatedlora}.

Despite this progress, two key challenges remain. First, standard vision transformers rely on dense attention, leading to high energy consumption. Second, PEFT-based continual learning has been studied almost exclusively in artificial neural networks \cite{sdlora,l2p,wang2022dualprompt, smith2023coda, cl-lora}, while \emph{spiking neural networks}, known for their event-driven and energy-efficient computation, remain largely unexplored. Sparse event-based transformers such as SpikFormer \cite{spikformer}, QKFormer \cite{zhou2024qkformer}, and SpikingFormer \cite{zhou2026spikingformer} replace dense attention operations with spike-based computation, significantly improving energy efficiency. For CIL on SpikFormer \cite{spikformer} trained from scratch, \citet{nagabhushana2026catformer} recently proposed a threshold-only adaptation strategy for only Class Incremental and Task Incremental learning. However, continual learning in sparse event-based transformers remains underexplored, particularly in rehearsal-free and pre-trained settings for Online Continual Learning (OCL).

In this work, we introduce \textbf{sLoTh}, a rehearsal-free continual learning framework for sparse event-based vision transformers. The integration of sLoTh with sparse event-based transformers enables an energy and memory-efficient continual learning framework. The proposed framework combines three mechanisms: \textbf{seLoRA}, a scalable-efficient low-rank attention adaptation module that enables efficient representational updates; \textbf{threshold modulation}, which dynamically adjusts neuronal firing thresholds to regulate intrinsic plasticity in spiking networks, and \textbf{multi-objective distillation and regularisation} that preserves previously learned knowledge without requiring data replay. Our approach enables scalable continual adaptation without network expansion \cite{dsdsnn} or exemplar buffers \cite{ni2025alade, der,s6mod}. Accordingly, we evaluate our method across TIL, CIL, and OCL settings with necessary algorithmic modifications. 

Our contributions in this work are:
(1) We provide the first systematic study of PEFT methods for pretrained sparse event-based vision transformers in continual learning.
(2) We introduce \textbf{sLoTh}, a data rehearsal-free continual learning framework that combines shared \textbf{threshold modulation} with scalable, efficient low-rank attention adaptation (\textbf{seLoRA}) to enable stable adaptation without replay buffers or architectural expansion.
(3) Unlike prior LoRA-based continual learning approaches \cite{sdlora,liang2024inflora,cl-lora} that rely on task-specific modules or parameter growth, sLoTh performs \textbf{shared adaptation} optimised through a \textbf{multi-objective regularisation} scheme while preserving intrinsic plasticity in spiking networks.
(4) The proposed sLoTh framework in the CIL setup outperforms prior work, especially for high task counts ($ \geq20$) on CIFAR100 and Imagenet-R under a memory and energy budget.
(5) Our approach updates \textbf{less than 1\% of model parameters} while leveraging the efficiency of sparse event-driven transformers, which achieve \textbf{6.5$\times$ lower energy consumption} than conventional dense vision transformer architectures.
(6) It also demonstrates strong performance across challenging \textbf{online continual learning} benchmarks, where sLoTh surpasses previous methods despite not storing past samples.

\section{Related Work}

\subsection{Continual Learning}

Continual learning (CL) enables models to learn from data that arrives sequentially while mitigating catastrophic forgetting. Existing approaches are commonly grouped into regularisation-based, rehearsal-based, and architecture-based methods \cite{threeparadiagms}. Regularisation-based approaches constrain parameter updates through importance-weighted penalties such as EWC \cite{ewc}, R-EWC \cite{rewc}, and Synaptic Intelligence \cite{si}. Rehearsal-based methods approximate the joint training distribution by replaying stored samples or features, including Experience Replay (ER) \cite{rolnick2019experience}, GEM \cite{GEM}, A-GEM \cite{AGEM}, and Dark Experience Replay (DER)\cite{der}. For OCL, replay-based strategies remain dominant \cite{rolnick2019experience,der}. Architecture-based approaches expand or gate model components to isolate task-specific parameters, though most were developed for convolutional networks and do not directly address parameter-efficient adaptation of pretrained transformers.

\subsection{Parameter-Efficient Fine-Tuning for Continual Adaptation}

Parameter-Efficient Fine-Tuning (PEFT) enables adaptation of large pretrained transformers by updating only a small subset of parameters while keeping the backbone frozen \cite{ptm-survey}. Existing methods include prompt-based approaches such as L2P \cite{l2p}, Coda-Prompt \cite{smith2023coda}, and DualPrompt \cite{wang2022dualprompt}; adapter-based modules inserted within transformer layers; and low-rank adaptation methods such as LoRA and its continual variants, including SD-LoRA \cite{sdlora} and Online LoRA \cite{wei2025onlinelora}. 

% While effective for dense transformers, these methods often rely on task-specific modules, limiting their applicability in efficient online continual learning settings.

% \subsection{Online Continual Learning}
% Online Continual Learning (OCL) considers streaming settings in which each sample is typically observed only once and task boundaries are unknown \cite{aljundi2019taskfree, ye2024onlinetaskfree}. Replay-based strategies remain dominant, where methods such as Experience Replay (ER) \cite{rolnick2019experience} and Dark Experience Replay (DER) \cite{der} maintain memory buffers to approximate the joint training distribution. However, replay introduces memory overhead and may be unsuitable in privacy-sensitive or resource-constrained scenarios, motivating the development of rehearsal-free alternatives for stable online adaptation.

\subsection{Continual Learning in Sparse Event-Based Networks}

Spiking Neural Networks (SNNs) enable event-driven computation through membrane potential accumulation and threshold-based firing, offering improved energy efficiency compared to conventional neural networks. Recent work extends transformer architectures into this domain through spike-based attention mechanisms, as demonstrated by models such as Spikformer \cite{spikformer}, SpikingFormer \cite{zhou2026spikingformer}, and QKFormer \cite{zhou2024qkformer}. However, these studies don't focus on continual learning. Continual learning in spiking networks typically relies on CNN or MLP backbones \cite{dsdsnn, SASNN}, sometimes with rehearsal buffers \cite{ni2025alade} or via task-specific dynamic thresholds for Spikformer \cite{spikformer} trained from scratch \cite{nagabhushana2026catformer}. However, no prior works analyse parameter-efficient fine-tuning (PEFT) methods for continual learning using pretrained sparse event-based transformers.

% Parameter-efficient continual adaptation in pretrained spiking transformers, although CATFormer \cite{nagabhushana2026catformer} shows some initial developments, particularly in the usage of dynamic thresholds for continual learning, but doesn't work for online continual learning due to task-dependent thresholds chosen via MLP.

\section{Methodology}
% \begin{algorithm}[htbp]
% \caption{sLoTh: Online Continual Learning with Sparse Event-Based Transformers}
% \label{alg:sloth}
% \begin{algorithmic}[1]
% \STATE \textbf{Input:} Pretrained backbone $f_\theta$, adaptation parameters $\psi$, threshold $\gamma$
% \STATE Initialise EMA loss $\bar{\mathcal{L}}*0$, teacher parameters $\psi^\ast$
% \FOR{each batch $(x_b, y_b)$ in data stream}
% \STATE $z \leftarrow f*{\theta,\psi}(x_b)$ \hfill \COMMENT{feature extraction}
% \STATE Compute logits using cosine classifier
% \STATE Compute loss $\mathcal{L}*b = \mathcal{L}*{CE} + \mathcal{L}_{stab}$
% \STATE Update EMA loss $\bar{\mathcal{L}}*b = (1-\alpha)\bar{\mathcal{L}}*{b-1} + \alpha\mathcal{L}_b$
% \STATE $r_b \leftarrow \mathcal{L}_b / \bar{\mathcal{L}}_b$
% \IF{$r_b > \gamma$} \hfill \COMMENT{novelty detection}
% \STATE Save current parameters $\psi^\ast \leftarrow \psi$
% \STATE Register new class prototypes
% \ENDIF
% \STATE Update adaptation parameters $\psi$ via gradient descent
% \ENDFOR
% \STATE \textbf{Inference:} classify using nearest class mean (NCM)
% \end{algorithmic}
% \end{algorithm}
We propose a rehearsal-free continual learning framework for sparse event-based transformers, designed for energy-efficient adaptation in resource-constrained environments. By restricting plasticity to lightweight adaptation modules while keeping the pretrained backbone frozen, the proposed approach enables continual adaptation with minimal parameter updates and computational overhead. Our primary focus is on the Class-Incremental Learning (CIL) and Task-Incremental Learning (TIL) settings. We additionally extend the framework to the more challenging Online Continual Learning (OCL) setting through minor modifications, enabling adaptation under single-pass streaming without explicit task boundaries.
% Our approach restricts plasticity to lightweight, biologically-inspired modules while keeping the pretrained backbone frozen, drawing on neuromodulatory mechanisms in biological systems where stable synaptic structures are modulated by low-dimensional control signals and intrinsic excitability adjustments.
\subsection{Problem Formulation}

We consider a sequence of tasks $\{\mathcal{D}_t\}_{t=1}^T$, where each task $\mathcal{D}_t=\{(x_i,y_i)\}_{i=1}^{N_t}$ introduces a disjoint label space $\mathcal{C}_t$. The cumulative label space after task $t$ is $\mathcal{C}_{\leq t}=\bigcup_{k=1}^{t}\mathcal{C}_k$. We evaluate \textbf{Full sLoTh}, comprising seLoRA and threshold modulation, under three continual learning protocols i.e CIL, TIL and OCL. Refer to appendix \ref{appendix: problem_formulation} for detailed problem formulation.

% In the OCL setting, task transitions are implicitly inferred using the proposed novelty detection mechanism (Section~\ref{sec:novelty}). When a significant distribution shift is detected, the model treats it as a pseudo-task boundary, triggering teacher snapshotting and classifier expansion. We emphasise that this mechanism is intended as a lightweight and practical approach for task-free adaptation under current OCL benchmarks, while more sophisticated distribution-shift estimation strategies remain an important direction for future work.

\subsection{Model Plasticity}
Let $f_{\theta}(\cdot)$ denote a pretrained sparse event-based transformer with frozen backbone parameters $\theta$. To avoid catastrophic forgetting caused by representation drift, we restrict plasticity to lightweight adaptation parameters while keeping the backbone fixed. Low-rank adaptation (LoRA) provides a parameter-efficient strategy for continual learning \cite{hu2022lora, liang2024inflora, sdlora, wei2025onlinelora}, but many approaches introduce task-specific modules \cite{adaptercil, liang2025gatedlora} or modify attention projections, leading to parameter growth and feature drift in single-pass continual learning. To address this, we introduce \emph{channel-wise excitability modulation}, which adjusts neuronal firing thresholds instead of synaptic weights to adapt feature responses while preserving the pretrained representation. For multi-epoch settings such as CIL and TIL, we additionally enable scalable, efficient low-rank attention updates to increase adaptation capacity.

\subsubsection{Channel-wise Excitability Modulation}

In sparse event-based (spiking) networks, neurons emit spikes when their membrane potential exceeds a firing threshold. Modulating this threshold alters neuronal excitability without changing synaptic weights, a mechanism widely observed in biological neural systems where neuromodulators regulate neuronal gain and firing thresholds \cite{hammouamri2022mitigating,ding2022biologically}.

We introduce learnable \emph{per-channel} threshold offsets that modulate neuronal excitability:

\begin{equation}
v_{\text{th}}(c) = v_{\text{th}}^{\text{base}} + \delta_c,
\end{equation}

where $v_{\text{th}}^{\text{base}}$ is the pretrained base threshold and ${\delta_c}$ is learnable offsets for each feature channel. Each parameter $\delta_c$ controls the excitability of an entire feature map across spatial locations, allowing the model to selectively amplify or suppress specific channels in response to new tasks. Unlike weight-space adaptation methods such as LoRA \cite{hu2022lora}, threshold modulation adjusts intrinsic neuronal excitability rather than synaptic connectivity. Because this mechanism modulates neuronal excitability without directly altering pretrained feature projections, it introduces less representational disruption than weight-space adaptation methods. This behaviour is particularly beneficial for continual learning approaches that rely on stable feature-space prototypes.
% For each attention projection matrix $W \in \mathbb{R}^{d \times d}$ (query, key, value), we introduce a low-rank residual update:
% \begin{equation}
% W' = W + \Delta W, \qquad \Delta W = BA,
% \end{equation}
% where $B \in \mathbb{R}^{d \times r}$ and $A \in \mathbb{R}^{r \times d}$ with $r \ll d$. The matrices $A$ and $B$ form the learnable parameters $\phi = {A_i, B_i}_i$.

\subsubsection{Low-Rank Attention Modulation}
\label{sec:lora}
We incorporate low-rank attention modulation inspired by LoRA \cite{hu2022lora}. The low matrices $A$ and $B$ form the learnable parameters $\phi = {A_i, B_i}_i$.  In our design, we refer to this adaptation scheme as \textbf{seLoRA} (\emph{Scalable-Efficient LoRA}), where a single shared low-rank module provides parameter-efficient plasticity without allocating task-specific adapters and is optimised via a stabilised training objective (see \ref{sec: stability}). Several continual learning methods employ LoRA-style adapters that expand with the number of tasks or rely on explicit task boundaries \cite{wei2025onlinelora,sdlora, liang2025gatedlora}. These methods make it difficult to apply in an OCL setting, as we have no task boundaries. However, under the OCL regime, jointly updating seLoRA and threshold modulation leads to unstable optimisation. While each mechanism independently provides sufficient plasticity ($\psi={\phi}$ or $\psi={\delta}$), their combined updates amplify representation drift when the model observes each sample only once. In contrast, CIL and TIL allow multiple passes over each task dataset, which stabilises optimisation and enables both mechanisms to be used jointly ($\psi={\phi,\delta}$).

\subsection{Stabilized Training Objective}
\label{sec: stability}
The training objective balances \emph{adaptation} to new classes with \emph{stabilisation} of previously learned representations: 
\begin{equation}
\mathcal{L} = {\mathcal{L}_{\text{CE}}} \;+
% {\mathcal{L}_{\text{stab}}}.\ \ \ where, \mathcal{L}_{\text{stab}} =
\lambda_{\text{KL}}\,\mathcal{L}_{\text{KL}} + \lambda_{\text{local}}\,\mathcal{L}_{\text{local}} + \lambda_{\text{global}}\,\mathcal{L}_{\text{global}}
\end{equation}
% \begin{equation}
%     \mathcal{L}_{\text{stab}} = \lambda_{\text{KL}}\,\mathcal{L}_{\text{KL}} + \lambda_{\text{local}}\,\mathcal{L}_{\text{local}} + \lambda_{\text{global}}\,\mathcal{L}_{\text{global}}.
% \end{equation}
where $\mathcal{L}_{\text{CE}}$ refers to the classifier trained using cross-entropy over the current task. The stability term comprises teacher-student logit distillation and local and global parameter anchors. Further details are provided in Appendix \ref{appendix:stabilised training objective} and analysis in section \ref{sec: loss_analysis}.

% -----------------------------------------------------

% % -----------------------------------------------------
\subsection{Classifier and Prototype Inference}

Given a feature representation $z = f_{\theta,\psi}(x) \in \mathbb{R}^d$, logits are computed using a cosine classifier $\ell_c(x) =
s \cdot
\frac{z}{|z|_2}
\cdot
\frac{w_c}{|w_c|_2}$ where $w_c \in \mathbb{R}^d$ denotes the classifier weight for class $c$ and $s$ is a fixed temperature. After completing task $t$, the classifier columns corresponding to classes ${w}_{c \in \mathcal{C}_t}$ are frozen, and only the weights of newly introduced classes are updated in subsequent tasks. While the cosine classifier is used during training, we employ Nearest Class Mean (NCM) classification at inference time to reduce recency bias. This method is widely used in the literature and has been employed in prior work such as  \cite{rebuffi2017icarl,l2p,wang2022dualprompt,lucir}. After task $t$, a prototype for each class is computed as $\mu_c$. During inference, predictions are obtained by nearest prototype matching as $\hat{y}$.
\begin{equation}
\mu_c =
\frac{1}{|{i : y_i=c}|}
\sum_{y_i=c}
f_{\theta,\psi}(x_i). \ \ \  and  \ \ \hat{y} =
\arg\max_{c \in \mathcal{C}_{\le t}}
\cos\!\left(
f_{\theta,\psi}(x),\, \mu_c
\right).
\end{equation}

% \begin{equation}
% \mu_c =
% \sum_{y_i=c}
% f_{\theta,\psi}(x_i)/|{i : y_i=c|}.
% \end{equation}

% In an OCL setting, storing all past samples is infeasible. 
% Therefore, prototypes are updated incrementally using a running mean:
% \begin{equation}
% \mu_c^{(t)} = \mu_c^{(t-1)} + \frac{1}{n_c^{(t)}} \left( f_{\theta,\psi}(x_t) - \mu_c^{(t-1)} \right),
% \end{equation}
% where $n_c^{(t)}$ is the number of observed samples for class $c$ up to step $t$.

% During inference, predictions are obtained by nearest prototype matching: 
% ---------------------------------------------
\subsection{Online Novelty Detection in Event-Based Transformers}
\label{sec:novelty}

In the OCL setting, task boundaries are not provided to the model \cite{aljundi2019taskfree, ye2024onlinetaskfree}. The data, therefore, arrive as a continuous stream, and distribution shifts may indicate the introduction of previously unseen classes. Detecting such shifts is closely related to concept drift detection in streaming learning systems \cite{oodsurvey}. We detect distribution changes online using the dynamics of the training loss. Specifically, we maintain an exponential moving average (EMA) of the cross-entropy loss by $\bar{\mathcal{L}}_b = (1-\alpha)\bar{\mathcal{L}}_{b-1}
+
\alpha\mathcal{L}_b$ where $\mathcal{L}_b$ is the loss at batch $b$ and $\alpha$ controls the smoothing factor. At each step, we compute the loss ratio $r_b = {\mathcal{L}_b}/{\bar{\mathcal{L}}_b}.$ A novelty trigger fires when $r_b > \gamma$, where $\gamma$ is a predefined threshold hyperparameter (e.g., $\gamma=2$ indicates that the loss has doubled relative to the running average). Such abrupt increases in loss may indicate distribution shifts or the appearance of previously unseen classes \cite{oodsurvey}. 
We emphasise that this mechanism is intended as a starting point for task-free adaptation in sparse event-based transformers rather than a fully robust drift-detection strategy. Unlike several ANN-based OCL methods that rely on replay buffers, clustering, or explicit task-boundary supervision, our approach prioritises minimal memory overhead and computational simplicity, making it suitable for resource-constrained streaming settings. When a trigger is activated, the model snapshots the current adaptation parameters $\psi$ as a new teacher for the local anchor and registers newly observed classes. More advanced distribution-shift detection mechanisms may further improve robustness in future work.

% \subsection{Extension to CIL and TIL}
% \label{sec:cil_extension}
% The proposed OCL framework extends naturally to CIL and TIL with two modifications. First, since tasks are trained over multiple epochs, we allow seLoRA modulation (\ref{sec:lora}) in addition to thresholds, resulting in adaptation parameters $\psi={\phi,\delta}$. Second, as task boundaries are known, novelty detection (\ref{sec:novelty}) is replaced with deterministic teacher snapshots taken at each task transition.

\subsection{Extension to Online Continual Learning}
\label{sec:cil_extension}

The proposed framework is primarily designed for the CIL and TIL settings, where task data are available over multiple training epochs and task transitions are explicitly defined. In these settings, both threshold modulation and seLoRA adaptation are jointly optimised, resulting in adaptation parameters $\psi=\{\phi,\delta\}$. Teacher snapshots are deterministically updated at each task transition to preserve previously learned knowledge.
We also extend the framework to the more challenging OCL setting by introducing two modifications. First, since data are observed only once in a streaming manner and task boundaries are unavailable, deterministic teacher updates are replaced by the proposed novelty-detection mechanism (Section~\ref{sec:novelty}), which triggers adaptive teacher snapshotting based on distribution shifts.

% ----------------------------------------------------------------------
\section{Experiments}
We evaluate sLoTh, a parameter-efficient continual learning framework designed for sparse event-based transformers. Our experiments address two key questions - (1)\textit{Which PEFT mechanisms are compatible with sparse event-based transformers under resource constraints?} (2) \textit{Can rehearsal-free event-based adaptation compete with replay-based continual learning methods with generalisation across CIL and OCL settings?}
% \begin{enumerate}
%     \item Which PEFT mechanisms are compatible with sparse event-based transformers under resource constraints?
%     \item Can rehearsal-free event-based adaptation compete with replay-based continual learning methods with generalisation across CIL and OCL settings?
%     % \item Does the proposed framework generalise across CIL and OCL settings?
% \end{enumerate}

\subsection{Experimental Setup}

% All experiments are conducted using pretrained sparse event-based transformer backbones.
We primarily use QKFormer \cite{zhou2024qkformer}, which replaces quadratic attention with computationally efficient Q-K attention, improving memory efficiency and scalability while maintaining strong accuracy. We additionally evaluate on SpikingFormer, given in table \ref{tab:cil_results_extended} an ANN ViT-Base models to assess transferability across different transformer architectures. These results suggest that the proposed framework generalises with minor architectural modifications and limited hyperparameter tuning. We report two standard continual learning metrics: \textbf{Average Accuracy (AA)} and \textbf{Average Forgetting (F)}. Let $a_{t,i}$ denote the accuracy on task $i$ after learning task $t$, with $T$ total tasks. The final average accuracy is defined as
$\text{AA} = \frac{1}{T} \sum_{i=1}^{T} a_{T,i},$
while forgetting task $i$ is
$F_i = \max_{t \in \{1,\dots,T\}} \{a_{t,i} - a_{T,i}\}.$
The average forgetting is then
$F = \frac{1}{T-1} \sum_{i=1}^{T-1} F_i.$
AA measures retained performance across tasks, whereas forgetting quantifies the degradation of stability during continual adaptation.
\subsection{Class Incremental Learning}

% We seek to identify class incremental learning (CIL) methods for real-world on-device learning scenarios, where performance, energy, and memory efficiency are equally important.  Hence, 
We evaluate our sLoTh on sparse event-based transformers, which are more energy efficient than ANN-based transformers (refer to empirical validation in Table \ref{tab: split_cifar100_vit_ocl}). Moreover, previous studies have well established that event-based transformers perform similarly to ANN-based transformers \cite{vit}. From Table \ref{tab:cil_results}, we observe that while several ViT-based PEFT approaches perform strongly under a small number of task settings (for example, SD-LoRA achieves \textbf{88.01\%} for 10 tasks), their performance either degrades significantly or becomes computationally infeasible in constrained environments as the number of tasks increases.  In contrast, Full sLoTh (with seLoRA and a learnable threshold) demonstrates improved scalability as the task sequence becomes more fragmented, i.e., with a large number of tasks. Although its performance is slightly lower under coarse splits (\textbf{81.23\%} across 10 tasks), it remains significantly energy-efficient due to its sparse event-based design. It improves with finer task granularity, reaching \textbf{84.65\%} for 20 tasks and \textbf{87.22\%} for 50 tasks, outperforming all compared methods. We would like to refer the reader to Table \ref{tab:ablation_transition} for further analysis on energy and parameters. To further establish the efficacy of sLoTh, we next evaluate performance on a higher-resolution dataset, \textbf{ImageNet-R}.
%  Unlike the other recent methods, the model size of Full sLoTh is significantly smaller, thereby potentially reducing device memory requirements.

% It is a benchmark containing stylised, artistic, and naturally corrupted variants of ImageNet images. Unlike standard ImageNet, these images exhibit significant variations in texture, colour, and artistic style, making the dataset a challenging benchmark for evaluating robustness to distribution shifts.

% \ref{tab:cil_results} evaluates sLoTh in the CIL setting on CIFAR-100 under different task granularities. 
\begin{table*}[htbp]
\centering
\caption{CIFAR100 CIL comparison between pretrained ViT with PEFT Continual Learning methods v/s pretrained QKFormer on \textbf{Full sLoTh} for different task granularity, i.e., $T$ represents the number of tasks. Parameters are in millions. \textbf{First} and \underline{Second} best models are highlighted.}
\resizebox{\textwidth}{!}{
\begin{tabular}{lccccc}
\hline
\textbf{Methods} & \textbf{T=10} & \textbf{T=20} & \textbf{T=50} & \textbf{Total Param(M)} & \textbf{$\#$updates(M)} \\
\hline
L2P \cite{l2p}& 83.18$\pm$1.20 & 79.51$\pm$0.67 & 67.95$\pm$2.12 & 172 & \textbf{0.12} \\
DualPrompt \cite{wang2022dualprompt}& 81.48$\pm$0.86 & 80.44$\pm$1.38 & 72.5$\pm$1.08 & 172 & 0.86 \\
CODA-Prompt \cite{smith2023coda} & 86.31$\pm$0.12 & \underline{81.36$\pm$0.88} & \underline{73.77$\pm$0.98} & 172 & 4.6 \\
InfLoRA \cite{liang2024inflora}& \underline{86.75$\pm$0.35} & 80.97$\pm$0.74 & 70.68$\pm$1.26 & 172 & 0.51 \\
SD-LoRA \cite{sdlora}& \textbf{88.01$\pm$0.31} & OOM & OOM & 172 & \underline{0.39} \\
\textbf{Full sLoTh} & 81.23$\pm$1.74 & \textbf{84.65$\pm$1.46} & \textbf{87.22$\pm$1.59} & \textbf{64.32} & {0.44} \\
\hline
\end{tabular}}
\label{tab:cil_results}
\end{table*}
% =====================================================
% \subsection{Robustness under Domain Shift}
From Table \ref{tab:imagenetr}, we observe that while several ViT-based PEFT approaches achieve strong performance under coarse task splits, their accuracy degrades as the number of tasks increases. For example, SD-LoRA achieves \textbf{79.15\%} accuracy for $T=5$ tasks but drops to \textbf{75.26\%} when $T=20$. In contrast, Full sLoTh demonstrates improved scalability as the number of tasks grows. Although its performance is slightly lower under coarse splits ($\textbf{71.61\%}$ for $T=5$), it remains significantly energy- and memory-efficient. Moreover, with finer task granularity, Full sLoTh reaches \textbf{77.10\%} for $T=20$, outperforming all compared methods. These observations suggest that spiking temporal dynamics, combined with lightweight, parameter-efficient adaptation, provide an efficient CIL method even when the number of incremental updates increases.
% This trend suggests that combining spiking temporal dynamics with lightweight, parameter-efficient adaptation improves robustness to distribution shifts, particularly as the continual learning problem becomes more fragmented.
% \ref{tab:imagenetr} evaluates robustness under distribution shift using ImageNet-R, a benchmark containing stylised, artistic, and naturally corrupted variants of ImageNet images. Unlike standard ImageNet, these images exhibit significant variations in texture, colour, and artistic style, making the dataset a challenging benchmark for evaluating robustness to distribution shifts.

\begin{table*}[htbp]
\centering
\caption{CIL under domain shift on ImageNet-R. $T$ denotes the number of task splits. Accuracy is reported as mean $\pm$ std over 3 runs. \textbf{First} and \underline{Second} best models are highlighted.}
\begin{tabular}{lccc}
\hline
Method & Imagenet-R ($T=5$)& Imagenet-R ($T=10$)& Imagenet-R ($T=20$) \\
\hline
Full Fine tuning & 64.92$\pm$0.87 & 60.57$\pm$1.06 & 49.95$\pm$1.31 \\
L2P \cite{l2p}& 73.04$\pm$0.71 & 71.26$\pm$0.44 & 68.97$\pm$0.51 \\
DualPrompt \cite{wang2022dualprompt}& 69.99$\pm$0.57 & 68.22$\pm$0.20 & 65.23$\pm$0.45 \\
CODA-Prompt \cite{smith2023coda} & 76.63$\pm$0.27 & 74.05$\pm$0.41 & 69.38$\pm$0.33 \\
InfLoRA \cite{liang2024inflora}& \underline{76.95$\pm$0.23} & 74.75$\pm$0.64 & 69.89$\pm$0.56 \\
SD-LoRA \cite{sdlora}& \textbf{79.15$\pm$0.20} & \textbf{77.34$\pm$0.35} & \underline{75.26$\pm$0.37} \\
\hline
\textbf{Full sLoTh } & 71.61$\pm$0.92 & \underline{76.17$\pm$1.71} & \textbf{77.10$\pm$1.13} \\
\hline
\end{tabular}
\label{tab:imagenetr}
\end{table*}

% =====================================================
% \section{Ablation}
% \label{sec: ablation}

\subsection{Extension to Online Continual Learning}
\label{sec: ocl}
% We further investigate whether the proposed parameter-efficient adaptation framework remains effective under the stricter Online Continual Learning (OCL) setting, where data arrive in a single-pass stream without explicit task boundaries. We follow the strict OCL protocol in which each sample is observed only once in a \textbf{single-pass stream}, and task boundaries are not provided during training. Unlike most prior approaches, \textbf{sLoTh does not maintain a replay buffer}. Instead, sLoTh maintains only a lightweight \textit{per-class prototype vector}. Unlike replay-based methods that store thousands of raw samples, this representation stores only class-level statistics, resulting in significantly lower memory overhead while still preserving representative information for decision boundaries.

We further evaluate sLoTh under the stricter Online Continual Learning (OCL) setting, where data arrive in a single-pass stream and explicit task boundaries are unavailable during training. Unlike many existing OCL approaches, sLoTh does not rely on replay buffers or stored training samples. Instead, the framework maintains only lightweight per-class prototype statistics, substantially reducing memory overhead while preserving representative information for decision boundaries in streaming environments. Consistent with the CIL setting, we focus here on energy-efficient OCL using sparse event-based transformers. Table~\ref{tab:ocl_main} compares sLoTh against several replay-based OCL methods on CIFAR-100 ($T=10$) and Tiny-ImageNet ($T=100$). Unlike these baselines, which rely on replay buffers containing up to \textbf{5k} and \textbf{10k} stored samples, respectively, sLoTh operates without replay memory and maintains only lightweight prototype statistics. Despite these constraints, Full sLoTh achieves \textbf{55.12\%} accuracy with \textbf{12.79} forgetting on CIFAR-100, closely matching the replay-based MOE-MOSE baseline (\textbf{55.62\%}). These results suggest that structured intrinsic plasticity together with lightweight attention modulation can effectively mitigate catastrophic forgetting even under strict streaming conditions.

\begin{table}[htbp]
\centering
\caption{Online Continual Learning comparison under single-pass streaming. Replay-based results are taken from S6MOD under the largest buffer sizes (CIFAR100: M=5k, Tiny-ImageNet: M=10k).\textbf{ sLoTh operates without any replay buffer} has only per-class mean prototype. Accuracy is reported as mean $\pm$ std over 3 runs. \textbf{First} and \underline{Second} best models are highlighted.}
% \resizebox{\textwidth}{!}{
\begin{tabular}{lcccc}
\hline
& \multicolumn{2}{c}{\textbf{CIFAR100 (T=10)}} 
& \multicolumn{2}{c}{\textbf{Tiny-ImageNet (T=100)}}\\
\textbf{Method} 
& Avg Acc
& Forgetting
& Avg Acc
& Forgetting \\
\hline 
ER \cite{rolnick2019experience}& 39.41$\pm$1.81 & 13.29 & 24.71$\pm$2.52 & 40.77 \\
OCM \cite{OCM} & 42.22$\pm$1.06 & \textbf{3.76} & 31.94$\pm$1.19 & 15.92 \\
OnPro \cite{onpro}& 41.59$\pm$1.38 & 6.72 & 26.38$\pm$2.18 & 20.32 \\
OCM-CCLDC \cite{OCM-CCLDC}& 51.43$\pm$1.37 & \underline{3.99} & 39.25$\pm$0.88 & 15.56 \\
OnPro-CCLDC \cite{OCM-CCLDC} & 50.01$\pm$0.85 & 10.55 & 38.18$\pm$1.02 & 16.17 \\
MOSE \cite{MOSE} & 54.53$\pm$0.78 & 13.60 & 38.71$\pm$0.44 & 15.51 \\
MOE-MOSE \cite{MOSE}& 55.62$\pm$0.72 & 12.81 & 38.41$\pm$0.53 & \underline{13.94} \\
\hline
\textbf{Full sLoTh} & 55.12$\pm$1.18 & 12.79  & {42.90$\pm$2.56} & 14.63  \\
\textbf{sLoTh only with Thresholds} & \underline{60.25$\pm$0.98} & 11.36 & \underline{61.41$\pm$1.05} & \textbf{12.32} \\
\textbf{sLoTh only with seLoRA } &\textbf{ 63.88$\pm$1.03} & 13.85 & \textbf{65.12$\pm$2.52} & 15.16\\
% \textbf{sLoTh on spikingformer } & & & & \\
\hline
\end{tabular}
% }
\label{tab:ocl_main}
\end{table}

% As in the CIL scenario, we focus here on achieving energy-efficient OCL. Accordingly, in Table \ref{tab:ocl_main}, we evaluate sLoTh only on sparse event-based transformers. 
% Table \ref{tab:ocl_main} compares sLoTh against several state-of-the-art replay-based OCL methods on CIFAR-100 ($T=10$, where $T$ is the task count) and Tiny-ImageNet ($T=100$). These baselines rely on large replay buffers containing up to \textbf{5k samples for CIFAR100} and \textbf{10k samples for Tiny-ImageNet}, allowing them to revisit past data during training. Despite operating \textit{without any replay memory}, sLoTh achieves \textbf{55.12\%} accuracy and a forgetting score of \textbf{12.79} on CIFAR-100, closely matching the strongest replay-based baseline MOE-MOSE. Thus, demonstrating that structured intrinsic plasticity combined with lightweight attention modulation can effectively mitigate catastrophic forgetting even under strict streaming constraints.

\begin{figure}[htbp]
    \centering
    \includegraphics[width=1\textwidth]{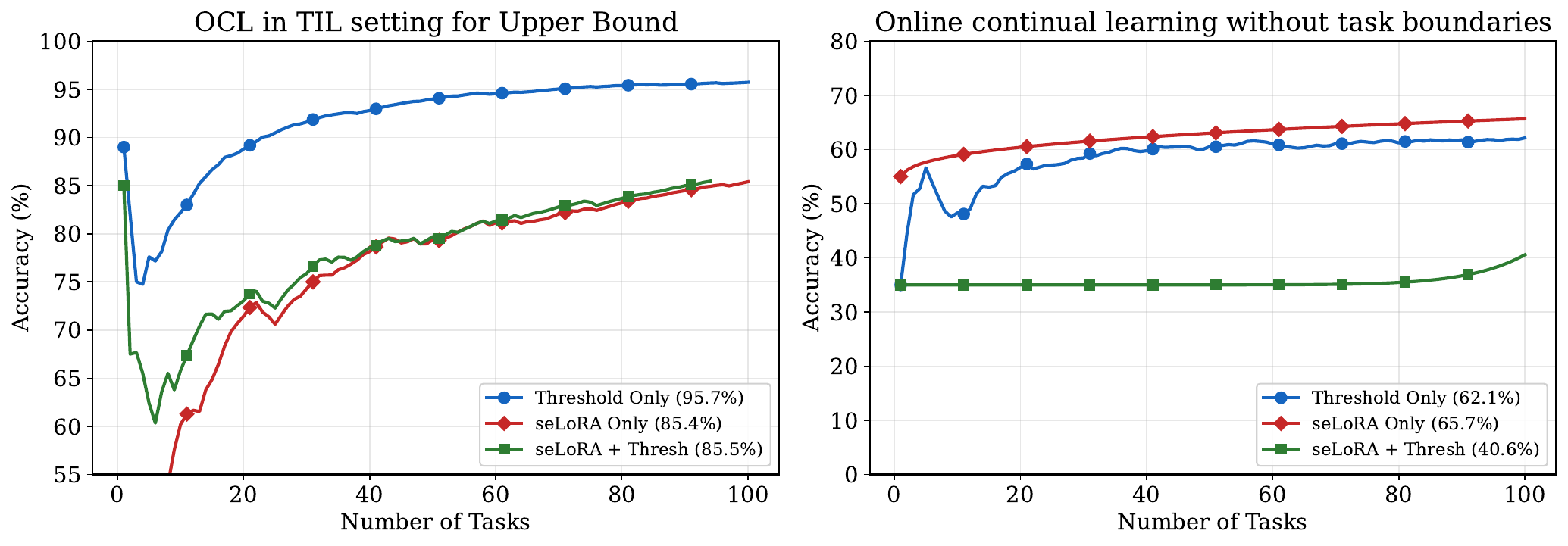}
    \caption{Performance gap between boundary-aware OCL v/s boundary-free OCL setting on Tiny-Imagenet. Here, sLoTh narrows the gap between these two settings compared to other methods without relying on replay.}
    \label{fig:OCL_tiny_im}
\end{figure}

% The advantage of sLoTh becomes more evident on the more challenging \textbf{Tiny-ImageNet} benchmark with \textbf{100 tasks}. Here, sLoTh achieves \textbf{42.90} accuracy, outperforming all replay-based baselines, including OCM (\textbf{31.94}), OnPro (\textbf{26.38}), and MOE-MOSE (\textbf{38.41}). This represents an improvement of approximately \textbf{+4.5\%} over the strongest replay-based baseline, suggesting that the proposed adaptation mechanism scales effectively as the number of tasks increases and the learning problem becomes more fragmented.
% Moreover, individual components of sLoTh, namely \textbf{sLoTh only with Thresholds} and \textbf{sLoTh only with seLoRA} adaptation, even outperform \textbf{Full sLoTh}, achieving \textbf{60.25} and \textbf{63.88} accuracy, respectively, on CIFAR100, and \textbf{61.41} and \textbf{65.12} on Tiny-ImageNet, respectively. These results highlight the strong role of intrinsic plasticity in stabilising spiking dynamics and the representational flexibility provided by scalable-efficient low-rank attention updates. However, these simplified variants achieve higher short-term accuracy in the strict streaming setting but introduce greater instability over longer task sequences. The full sLoTh framework, therefore, trades immediate plasticity for improved stability and retention, as further illustrated in Tables \ref{tab:ablation_transition} and \ref{tab:ocl_main}.
The effectiveness of sLoTh becomes more evident on \textbf{Tiny-ImageNet} benchmark with \textbf{100 tasks}. Full sLoTh achieves \textbf{42.90\%} accuracy, outperforming replay-based methods including OCM (\textbf{31.94\%}), OnPro (\textbf{26.38\%}), and MOE-MOSE (\textbf{38.41\%}), corresponding to an improvement of approximately \textbf{+4.5\%} over the strongest replay baseline. The simplified variants further improve short-term adaptation in the strict streaming setting. \textbf{sLoTh only with Thresholds} achieves \textbf{60.25\%} and \textbf{61.41\%} accuracy on CIFAR-100 and Tiny-ImageNet respectively, while \textbf{sLoTh only with seLoRA} achieves \textbf{63.88\%} and \textbf{65.12\%}. These observations suggest that stronger plasticity improves rapid adaptation under single-pass updates, whereas the combined Full sLoTh formulation provides a more stable trade-off between adaptation and long-term retention (Tables~\ref{tab:ablation_transition} and \ref{tab:ocl_main}).Figure~\ref{fig:OCL_tiny_im} illustrates the learning dynamics under a \textit{task-aware upper bound}, where task boundaries are known, and the realistic \textit{boundary-free online learning} setting. In the task-aware case, threshold-only adaptation achieves the highest performance (\textbf{95.7\%}), highlighting the effectiveness of intrinsic threshold plasticity when task transitions are explicitly available. When task boundaries are removed, performance decreases across all methods due to the difficulty of detecting distribution shifts online. In the boundary-free OCL setting, sLoTh only with seLoRA achieves \textbf{65.7\%}, while threshold-only adaptation achieves \textbf{62.1\%}. Notably, threshold-only adaptation requires at least $\textbf{10}\times$ fewer parameter updates than seLoRA-based variants. Overall, the combined Full sLoTh framework progressively narrows the gap between the task-aware and realistic streaming settings as more tasks are observed.

\begin{table}[htbp]
\centering
\caption{Comparison of OCL on Split-CIFAR-100 with 10 tasks with ViT-B/16 backbone(86 Million parameters)\cite{vit} v/s QKFormer(64 Million parameters)\cite{zhou2024qkformer} as backbone for our method. We consider the closest comparable model sizes for a fair comparison, based on parameter count in millions and energy consumption in mJ. \textbf{First} and \underline{Second} best models are highlighted.}
\begin{tabular}{lcccc}
\hline
 Methods & Avg Acc ($\uparrow$) & Forgetting ($\downarrow$) & \# Updates & \ Energy \ \\
\hline
AGEM \cite{AGEM} & 12.67 $\pm$ 1.87 & 82.51 $\pm$ 2.27 & Full & 254.84\\
ER \cite{rolnick2019experience} & 44.85 $\pm$ 1.83 & 44.67 $\pm$ 4.29 &  Full & 254.84 \\
EWC++ \cite{ewc++} & 10.61 $\pm$ 0.74 & 84.10 $\pm$ 1.11 & Full & 254.84 \\
MIR \cite{mir} & 48.36 $\pm$ 3.11 & 43.41 $\pm$ 1.02 &Full & 254.84 \\
PCR \cite{lin2023pcr} & 48.48 $\pm$ 0.15 & 46.23 $\pm$ 1.29 & Full& 254.84 \\
% DER++ \cite{der++} & 36.64 $\pm$ 6.11 & 56.94 $\pm$ 7.55 & 86M & 254.84 \\
LODE (DER++) \cite{LODE} & 44.29 $\pm$ 1.48 & 45.54 $\pm$ 3.32 & Full & 254.84 \\
EMA (DER++) \cite{ema} & 42.28 $\pm$ 4.36 & 55.59 $\pm$ 1.48 & Full & 254.84 \\
EMA (RAR) \cite{ema} & 47.10 $\pm$ 0.82 & 50.01 $\pm$ 0.35 & Full & 254.84 \\
Online-Lora \cite{wei2025onlinelora} & 49.40 $\pm$ 1.36 & 41.74 $\pm$ 2.58 & 0.623& 254.84 \\
\hline
Full sLoTh &{55.12 $\pm$ 1.18} & \underline{12.79 $\pm$ 2.28}  &{0.439} & \textbf{38.91} \\
sLoTh only thresholds &\underline{60.25 $\pm$ 0.98} & \textbf{ 11.36 $\pm$ 2.41}  & \textbf{0.037} & \textbf{38.91}\\
sLoTh only seLoRA &\textbf{63.88 $\pm$ 1.03} & 13.85 $\pm$ 3.97  & \underline{0.402} & \textbf{38.91}\\
% \textbf{sLoTh on spikingformer } & & & & \\
\hline
\end{tabular}
\label{tab: split_cifar100_vit_ocl}
\end{table}
Overall, these results demonstrate that the \textbf{sLoTh framework can match or surpass replay-based continual learning methods while remaining energy- and memory-efficient.} Similarly, from Table \ref{tab:imagenet100_extended} we observe that the Full sLoTh method outperforms prior state-of-the-art methods with ViT backbone on ImageNet-100 with 50 tasks by at least 1.76\%. Online-LoRA \cite{wei2025onlinelora} when transferred to pretrained \textbf{ QKFormer } backbone \cite{zhou2024qkformer} gives us 61.87, which is approximately equivalent to Full sLoTh's performance. Moreover, seLoRA is also applicable for ViT (55.68\%), and for such large task counts, seLoRA only achieves the best performance of 70\%. 

% By leveraging structured intrinsic plasticity together with lightweight attention modulation, which hardly is \textbf{1\% parameter} update (Tables \ref{tab: split_cifar100_vit_ocl}, \ref{tab:ablation_transition}) for the given model size and is \textbf{6.5$\times$} energy efficient than standard vision transformers \cite{zhou2024qkformer}. Hence, sLoTh provides an efficient and scalable solution for continual learning in streaming environments. 

% \begin{table}[htbp]
% \caption{Comparison of SOTA methods v/s sLoTh for Imagenet-100 with 50 task.}
% \centering
% \begin{tabular}{|l|c|c|c|c|c|}
% \hline
% \textbf{Method}  &   L2P \cite{l2p} & MVP \cite{mvp} & Online LoRA \cite{wei2025onlinelora}& Full sLoTh & Methods on QKFormer\\
% \hline
% \textbf{Accuracy (\%)} & 56.25 $\pm$ 0.45 &  
%  54.33 $\pm$ 3.05 & {60.38 $\pm$ 2.59} & \textbf{62.14 $\pm$ 1.3} & \\
% \hline
% \end{tabular}
% \label{tab:imagenet100}
% \end{table}

\begin{table}[htbp]
\centering
\caption{Comparison of SOTA methods on ImageNet-100 with 50 tasks for OCL.}
\label{tab:imagenet100_extended}

% \small
% \setlength{\tabcolsep}{2pt}
\resizebox{\textwidth}{!}{
\begin{tabular}{lccccccc}
\toprule

& \multicolumn{3}{c}{ViT} 
& \multicolumn{4}{c}{QKFormer} \\

\cmidrule(lr){2-4}
\cmidrule(lr){5-8}

Method
& L2P
& Online LoRA
& seLoRA
& Online LoRA
& Full sLoTh
& Threshold only
& seLoRA only \\

\midrule

Accuracy
& 56.25 $\pm$ 0.45
& 60.38 $\pm$ 2.59
& 55.68 $\pm$ 0.97
& 61.87 $\pm$ 1.63
& 62.14 $\pm$ 1.30
& 66.91 $\pm$ 1.56
& \textbf{70.00 $\pm$ 3.85} \\

\bottomrule
\end{tabular}}
\end{table}

% ===============================================

% \begin{table}[htbp]
% \centering
% \caption{Comparison of SOTA methods on ViT v/s sLoTh for Imagenet-100 with 50 task for OCL.}
% \label{tab:imagenet100_extended}

% \small
% \setlength{\tabcolsep}{4pt}

% \begin{tabular}{|c|cccccc|}
% \hline
% Method
% & L2P(ViT)
% & Online LoRA(ViT)
% & Online LoRA (QKFormer)
% & Full sLoTh (QKFormer)
% & Threshold only (QKFormer)
% & seLoRA only (QKFormer)
% &seLoRA only (ViT)\\
% \hline
% Accuracy
% & 56.25 $\pm$ 0.45 

% & 60.38 $\pm$ 2.59 
% &
% & 62.14 $\pm$ 1.30 
% & 66.91 $\pm$ 1.56 
% & \textbf{70.00 $\pm$ 3.85} \\
% \hline
% \end{tabular}
% \end{table}

% =====================================================
\subsection{Class Incremental Learning versus Online Learning}

Table \ref{tab:peft_feasibility} indicates PEFT in spiking networks can be mediated by seLoRA only, Threshold only or both (Full sLoTh) with varying classification accuracies. From Table \ref{tab:ablation_transition} we observe that, under CIL, where multiple epochs are available per task, the interaction between scalable-efficient low-rank modulation (seLoRA) and threshold adaptation becomes beneficial. seLoRA provides weight updates, while threshold modulation regulates neuronal excitability as new classes are introduced. Their combination significantly improves performance, achieving \textbf{81.23\%} accuracy for Full sLoTh. From Table \ref{tab:ablation_transition}, we also observe that for CIL, seLoRA only and Threshold only can also independently attain nearby performances of 72.63\% and 69.39\%, respectively. Whereas, during OCL, different variants (seLoRA only or Threshold only) yield higher immediate accuracy (\textbf{63.88\%} and \textbf{60.25\%} vs. \textbf{55.12\%} for Full sLoTh). This behaviour arises because the OCL setting observes each sample only once, favouring highly plastic adaptation mechanisms that can rapidly adjust model parameters. Here, seLoRA parameters are randomly initialised and must learn meaningful projection subspaces. In OCL settings, the additional constraints introduced by the full objective (Full sLoTh) can slow this adaptation, leading to slightly lower short-term accuracy. These results indicate that, while seLoRA-only or Threshold-only is beneficial for immediate adaptation in streaming settings, dual plasticity (Full sLoTh) becomes essential when learning must remain consistent across longer task sequences.

% To understand the interaction between scalable, efficient low-rank attention modulation and threshold adaptation, we compare their individual and combined effects under both OCL and CIL on CIFAR-100 (see \ref{tab:ablation_transition}).

\begin{table}[htbp]
\centering
\caption{Component interaction under CIL and OCL (CIFAR-100, 10 tasks).}
\begin{tabular}{lccc}
\hline
Variant & CIL Avg Acc  & OCL Avg Acc  & param updates \\
\hline
seLoRA only (w/o Threshold) & 72.63$\pm$1.24 & \textbf{63.88$\pm$1.03}  & 0.402\\
Threshold only (w/o seLoRA) & 69.39$\pm$1.01 & 60.25$\pm$0.98  & $\mathbf{0.037_{(\ 10x \downarrow)}}$\\
Full sLoTh (seLoRA + Threshold) & \textbf{81.23$\pm$1.74} & {55.12$\pm$1.18}  & 0.439 \\
\hline
\end{tabular}
\label{tab:ablation_transition}
\end{table}
\textbf{Threshold Modulation for Parameter Efficient Continual Learning: }Spiking thresholds are a unique attribute of spiking neural networks. From Table \ref{tab:peft_feasibility} we observe that fine-tuning only thresholds reaches an accuracy of 70.56\% compared to 80.89\% with LoRA. From Tables \ref{tab:imagenet100_extended} and \ref{tab:ablation_transition}, we notice that for the continual learning scenario,  threshold modulation is an effective parameter-efficient fine-tuning method (requiring atleast $10 \times$ lower parameter updates than LoRA). The effect is most pronounced in the OCL scenario, where Threshold only achieves 60.25\% and 66.71\% accuracies on CIFAR100 and Imagenet100, respectively, which is $\sim$3\% lower than seLoRA but significantly more efficient. 
% ==============================================================================

\subsection{Component Analysis of Loss Function}
\label{sec: loss_analysis}
Table \ref{tab: component_analysis} analyses the contribution of individual components in the proposed loss formulation. Removing the KL regularisation term ($\lambda_{kl}=0$) leads to a significant performance drop from \textbf{55.12\%} to \textbf{10.23\%}, indicating that distribution alignment plays a critical role in stabilising continual updates. 
\begin{table}[htbp]
\caption{Component analysis of Loss function on CIFAR-100 dataset in OCL setting }
\centering
\begin{tabular}{lcccc}
\hline
Components & Non-zeros & $\lambda_{kl}=0$ & $\lambda_{\text{local}}=0$ & $\lambda_{\text{global}}=0$ \\
\hline
Accuracy & \textbf{55.12 $\pm$ 1.18} & 10.23 $\pm$ 1.10  & 21.64$\pm$ 2.38& 20.11 $\pm$ 1.09   \\
\hline
\end{tabular}
\label{tab: component_analysis}
\end{table}
Similarly, disabling the local and global consistency terms results in substantial degradation, reducing accuracy to \textbf{21.64\%} and \textbf{20.11\%}, respectively. These terms encourage consistency between feature representations across updates, preventing representational drift during streaming adaptation. Overall, the results demonstrate that each component of the loss contributes to stable, continual learning, and the full objective is necessary.

% \section{Discussion}
% We here propose the \textbf{sLoTh} framework, which outperforms existing online continual learning methods. As sLoTh is optimal for sparse event-based transformers, it is 
% sary to maintain robust performance in the OCL setting.

\section{Discussion}
We propose \textbf{sLoTh} as a state-of-the-art data rehearsal-free continual learning framework. sLoTh is optimal for sparse event-based transformers, where threshold-based dynamics can even allow the model to learn with significantly smaller parameter updates without changing the underlying features.
Our results suggest that appropriately modified parameter-efficient plasticity mechanisms can outperform data-replay-based strategies in online continual learning. In contrast to conventional approaches that depend on memory buffers, sLoTh maintains stability by combining structured attention modulation with excitability control. This highlights the potential of understanding and leveraging the intrinsic neuronal properties for continual learning under strict memory and computational constraints. In addition, while the proposed novelty detection mechanism enables task-free adaptation in streaming settings, more advanced shift-detection strategies may further improve robustness in OCL, making it more scalable. Moreover, sLoTh is also optimal for CIL, especially for fine-grained tasks relevant to real-world scenarios. As part of this, we also observe that the plasticity modality of CIL is not directly transferable to OCL, and variants of sLoTh can address all these requirements.

% \bibliographystyle{plainnat}
% \bibliography{references}

%%%%%%%%%%%%%%%%%%%%%%%%%%%%%%%%%%%%%%%%%%%%%%%%%%%%%%%%%%%%
\section*{Appendix}
\appendix

% \section{Technical appendices and supplementary material}
% \begin{itemize}
%     \item ViT results --> rebuttal
%     \item extended methods
%     \item hyper param n stuff
%     \item exp setup
%     \item imagenet100 results
%     \item TIL
%     \item Decomposition for the thresholds and weights.
%     \item ResNet (spiking)
% \end{itemize}

\section{Extended Method Details}

This section provides additional methodology details for the \textbf{sLoTh} framework. 
% ------------------------------------------------------------
\subsection{Problem Formulation Details}
\label{appendix: problem_formulation}
We consider a sequence of tasks $\{\mathcal{D}_t\}_{t=1}^{T}$ where each task
$\mathcal{D}_t = \{(x_i,y_i)\}_{i=1}^{N_t}$ introduces a disjoint label set
$\mathcal{C}_t$ such that $\mathcal{C}_i \cap \mathcal{C}_j = \varnothing$ for $i \neq j$.
The cumulative label space after task $t$ is

\begin{equation}
\mathcal{C}_{\le t} = \bigcup_{k=1}^{t} \mathcal{C}_k.
\end{equation}

Given an input $x$, the model predicts

\begin{equation}
\hat{y} \in \mathcal{C}_{\le t}.
\end{equation}
 As part of our study, we evaluate \textbf{Full sLoTh}, which comprises two components, i.e seLoRA and Threshold modulation, under three protocols:\\
\textbf{OCL:}
Training data arrive as a stream $\{(x_{t,b}, y_{t,b})\}$ with each sample observed once and no task identity $t$ is provided.
The model must autonomously detect distribution shifts and preserve knowledge without explicit task boundaries.\\
\textbf{CIL:}
Given tasks $\{D_t\}_{t \in \mathcal{T}}$ with $\mathcal{C}_i \cap \mathcal{C}_j=\varnothing$ for $i\neq j$, the task identity is unknown at test time; predictions satisfy $\hat{y} \in \mathcal{C}_{\le t}=\bigcup_{i=1}^{t}\mathcal{C}_i$.\\
\textbf{TIL:}
Given tasks $\{D_t\}_{t \in \mathcal{T}}$ with $p(\mathcal{X}_i)\neq p(\mathcal{X}_j)$ and $\mathcal{C}_i \cap \mathcal{C}_j=\varnothing$ for $i\neq j$, the task identity $t$ is known at test time; predictions satisfy $\hat{y} \in \mathcal{C}_t$.

In the OCL setting, explicit task boundaries are not provided to the model. Instead, task transitions are implicitly inferred through the proposed novelty detection mechanism (Section~\ref{sec:novelty}). When a significant distribution shift is detected, the model treats this event as a pseudo-task boundary, triggering parameter snapshotting and classifier expansion. Therefore, while no ground-truth task identity is available, the framework internally constructs adaptive boundaries based on data-driven signals.

% ------------------------------------------------------------
\subsection{Plasticity Modules}

The backbone network is a pretrained, sparse event-based transformer
$f_{\theta}(\cdot)$ with frozen parameters $\theta$.
Plasticity is restricted to a small set of adaptation parameters
$\psi$.

\begin{equation}
z = f_{\theta,\psi}(x)
\end{equation} , where $z \in \mathbb{R}^d$ is the feature representation. Depending on the continual learning setting, we use:
\begin{equation}
\psi =
\begin{cases}
\{\delta\} & \text{in case of OCL}\\
\{\phi,\delta\} & \text{in case of CIL or TIL}
\end{cases}
\end{equation} where $\delta$ represents threshold modulation parameters and
$\phi$ denotes low-rank attention parameters.

% ------------------------------------------------------------
\subsubsection{Channel-wise Excitability Modulation}

In the Leaky-Integrate and Fire neuron, when the membrane potential exceeds
a threshold $v_{\text{th}}$ the neuron produces spikes. To make the thresholds learnable instead of utilising the intrinsic dynamic and backpropagating on the thresholds itself we introduce channel-wise offsets $\delta_c$:
\begin{equation}
v_{\text{th}}(c) = v_{\text{th}}^{base} + \delta_c
\end{equation}
where $c$ indexes feature channels. This mechanism helps in modulating neurons' excitability while keeping the underlying synaptic weights frozen. Hence, the feature geometry produced by the pretrained backbone
remains largely preserved.

% ------------------------------------------------------------
\subsubsection{Scalable-Efficient LoRA (seLoRA)}

Extending the experiments from OCL to CIL, we additionally apply low-rank attention modulation inspired by LoRA. For each attention projection matrix $W \in \mathbb{R}^{d \times d}$, a low-rank residual update is introduced:
\begin{equation}
W' = W + BA ,
\end{equation}
where $B \in \mathbb{R}^{d \times r}$ and $A \in \mathbb{R}^{r \times d}, \quad r \ll d$ and $\phi=\{A_i,B_i\}$ are the learnable parameters. Unlike task-specific LoRA modules proposed in prior work \cite{sdlora,liang2024inflora,cl-lora}, seLoRA uses a single shared low-rank module across the task sequence. This prevents parameter growth as the number of tasks increases.

% ------------------------------------------------------------
\subsection{Stabilized Training Objective}
\label{appendix:stabilised training objective}

The overall training objective is

\begin{equation}
\mathcal{L} =
\mathcal{L}_{CE}
+
\mathcal{L}_{stab}
\end{equation}
where
\begin{equation}
\mathcal{L}_{stab} =
\lambda_{KL}\mathcal{L}_{KL}
+
\lambda_{local}\mathcal{L}_{local}
+
\lambda_{global}\mathcal{L}_{global}.
\end{equation}

\paragraph{Cross-Entropy Loss}

The classifier is trained using cross-entropy over the current task
classes:

\begin{equation}
\mathcal{L}_{CE} =
CE(\ell_{\mathcal{C}_t}(x),y).
\end{equation}

\paragraph{Logit Distillation}

To preserve knowledge of previous classes, we distil predictions from
a frozen teacher model:

\begin{equation}
\mathcal{L}_{KL} =
\tau^2 KL
\left(
softmax(\ell^{s}/\tau)
\parallel
softmax(\ell^{*}/\tau)
\right)
\end{equation}

where $\tau$ is the temperature parameter.

\paragraph{Local Parameter Anchor}

To prevent abrupt parameter drift, we anchor adaptation parameters to
their previous snapshot:

\begin{equation}
\mathcal{L}_{local} =
||\psi - \psi^{*}||_2^2.
\end{equation}

\paragraph{Global Parameter Anchor}

To prevent long-term drift across many tasks, we regularise parameters
toward their initial state:

\begin{equation}
\mathcal{L}_{global} =
\frac{1}{|\psi|}
\sum_{p \in \psi}
(p - p^{(0)})^2.
\end{equation}

% ------------------------------------------------------------
\subsection{Prototype-Based Inference}

During training, a cosine classifier is used:

\begin{equation}
\ell_c(x) =
s \cdot
\frac{z}{||z||_2} \cdot
\frac{w_c}{||w_c||_2}
\end{equation} After training, predictions are made using Nearest Class Mean
classification. Class prototypes are computed as
\begin{equation}
\mu_c =
\frac{1}{N_c}
\sum_{y_i=c}
f_{\theta,\psi}(x_i).
\end{equation}

Inference is performed via cosine similarity:

\begin{equation}
\hat{y} =
\arg\max_{c \in \mathcal{C}_{\le t}}
\cos(f_{\theta,\psi}(x),\mu_c).
\end{equation}

% ------------------------------------------------------------
\subsection{Online Novelty Detection}

To detect distribution shifts in OCL, we monitor the training loss. We maintain an exponential moving average
\begin{equation}
\bar{\mathcal{L}}_b =
(1-\alpha)\bar{\mathcal{L}}_{b-1}
+
\alpha\mathcal{L}_b.
\end{equation}
A novelty trigger is activated when

\begin{equation}
r_b = \frac{\mathcal{L}_b}{\bar{\mathcal{L}}_b} > \gamma.
\end{equation}
When triggered, the model:
\begin{enumerate}
\item snapshots current adaptation parameters
\item registers newly discovered classes
\item updates the teacher model for distillation
\end{enumerate}

% ------------------------------------------------------------
\subsection{Full Training Procedure}

Algorithm~\ref{alg:sloth} summarises the full training framework for OCL. Instead of triggers, we have an exclusive snapshot captured for adaptation after every task in the case of CIL and TIL 

\begin{algorithm}[htbp]
\begin{algorithmic}[1]
\STATE Initialize pretrained backbone $f_{\theta}$ and adaptation parameters $\psi$
\STATE Initialize EMA loss $\bar{\mathcal{L}}_0$
\FOR{each batch $(x_b,y_b)$}
\STATE $z \leftarrow f_{\theta,\psi}(x_b)$
\STATE Compute loss $\mathcal{L}_b$
\STATE Update EMA $\bar{\mathcal{L}}_b$
\IF{$\mathcal{L}_b / \bar{\mathcal{L}}_b > \gamma$}
\STATE snapshot teacher parameters
\ENDIF
\STATE Update adaptation parameters $\psi$
\ENDFOR
\end{algorithmic}
\label{alg:sloth}
% \caption{sLoTh Training Procedure}
\end{algorithm}
\section{Datasets}

We conduct evaluations across different datasets, including CIFAR-100, Tiny-ImageNet, ImageNet-100, and ImageNet-R. For CIFAR-100, we use the standard dataset provided through the \texttt{torchvision} library. Tiny-ImageNet is obtained from the official repository.\footnote{\url{http://cs231n.stanford.edu/tiny-imagenet-200.zip}} ImageNet-100\footnote{\url{https://www.kaggle.com/datasets/ambityga/imagenet100}} and ImageNet-R\footnote{\url{https://github.com/hendrycks/imagenet-r.git}} are subsets of the ImageNet-1K dataset that provide a reduced-scale benchmark and a domain-shift benchmark. To simulate continual learning scenarios, each dataset is partitioned into disjoint class subsets that form incremental tasks. For every experiment, class labels are remapped to ensure consistent label ordering across tasks. To maintain reproducibility and ensure fair evaluation, all random operations, including class permutations and data shuffling, are controlled using fixed random seeds.

\section{Hyperparameter Settings}

Most hyperparameters follow the official QKFormer implementation.\footnote{\url{https://github.com/zhouchenlin2096/QKFormer.git}} In the online continual learning (OCL) setting, we additionally perform a small hyperparameter search over the novelty-detection threshold used to trigger teacher model snapshots. We observe that very small thresholds lead to frequent false triggers, which increase computational overhead but do not significantly affect final performance. In contrast, overly large thresholds delay the detection of distribution shifts, reducing the model's ability to effectively adapt to newly introduced classes. In practice, we select a threshold value that balances timely distribution-shift detection with stable training dynamics.

\label{appendix:hyperparams_tables}

\begin{table}[htbp]
\caption{Hyperparameters used across datasets and continual learning settings (OCL, CIL, and TIL). }
\centering
\begin{tabular}{lllll}
\hline
\textbf{Hyperparameter} & \textbf{CIFAR-100}  & \textbf{Imagenet-R} & \textbf{Imagenet-100} &\textbf{Tiny-IM} \\
\hline
Batch size                   & 128   & 64   & 64  & 128 \\
Time steps (T)               & 4   & 4   & 4  & 4  \\
Learning rate                & 1e-4 & 3e-4 & 3e-4  & 1e-4\\
Epochs        & 50   & 30   & 30  & 30\\ 
Optimizer                    & AdamW   & AdamW   & AdamW  & AdamW  \\
\hline
\end{tabular}

\label{table:hyperparams_tables}
\end{table}

Each experiment is conducted three times with different seeds on the same hyperparameters, using seeds 0, 1, and 2. The following tables describe the hyperparameters used for each dataset.

\section{Extended Results}

\subsection{PEFT for Sparse Event-Based Vision Transformers}
Before introducing sLoTh, we first evaluate which PEFT mechanisms are compatible with sparse event-based transformers. Table \ref{tab:peft_feasibility} reports the classification accuracy of different PEFT strategies applied to a QKFormer backbone on CIFAR-100. Adapter-based \cite{adapters}, i.e., tuning of small feed-forward networks acting as bottleneck layers, achieves the highest accuracy (\textbf{87.21\%}), indicating that additional learnable modules can effectively adapt feature representations. However, adapters introduce task-specific parameters and increase memory and inference overhead, making them less suitable for online continual learning scenarios with strict resource constraints.
\begin{table}[htbp]
\caption{PEFT feasibility on pretrained QKFormer evaluated on the CIFAR-100.}
\centering
\begin{tabular}{lcccc}
\hline
\textbf{Method} & Low-Rank Adaptation\cite{hu2022lora} & \ Adapters \cite{adapters}\ & Spiking Adapters & \ Thresholds \ \\
\hline
\textbf{Accuracy}
 & 80.89 & \textbf{87.21} & 79.82 & 70.56 \\
\hline
\end{tabular}
\label{tab:peft_feasibility}
\end{table}

Low-rank attention modulation using LoRA provides a more efficient alternative, achieving \textbf{80.89\%} accuracy while requiring only lightweight updates to attention projections. Threshold-only adaptation is the most parameter-efficient strategy, modifying spiking neuron firing thresholds without introducing additional weights, and still achieves \textbf{70.56\%} accuracy.

These results reveal a trade-off: adapters offer strong performance but incur architectural overhead, while threshold adaptation is extremely efficient but limited in representational flexibility. Motivated by this observation, we propose the \textbf{sLoTh framework}, which combines \textbf{low-rank attention modulation} with \textbf{threshold-based intrinsic plasticity} to achieve efficient adaptation without affecting inference efficiency.

\subsection{Task-Incremental Learning and Class Incremental Learning}
In addition to the online continual learning (OCL) and class-incremental learning (CIL) evaluations presented in the main paper, we report results under the Task-Incremental Learning (TIL) protocol. In this setting, task boundaries are available during both training and evaluation, making the problem comparatively easier than CIL and OCL.
\begin{table}[htbp]
\centering
\caption{Task-Incremental Learning results for sLoTh across different datasets. Accuracy reported as mean $\pm$ std over 3 runs.}
% \resizebox{\textwidth}{!}{
\begin{tabular}{lcccc}
\hline
Tasks & Tiny-ImageNet & CIFAR-100 & ImageNet-R & Imagenet-100\\
\hline
10 & 85.23 $\pm$ 2.94 & 87.38 $\pm$ 1.63 & 86.77 $\pm$ 1.08 & 89.33 $\pm$ 1.78\\
20 & 87.17 $\pm$ 1.38 & 86.84 $\pm$ 2.54 & 88.45 $\pm$ 2.79 & 90.92 $\pm$ 1.63\\
50 & 96.50 $\pm$ 1.25 & 91.20 $\pm$ 2.71 & 93.61 $\pm$ 1.81 & 94.5 $\pm$ 1.95\\
\hline
\end{tabular}
\label{tab:til_results}
\end{table}
Table~\ref{tab:til_results} reports the performance of the proposed sLoTh framework on Tiny-ImageNet, CIFAR-100, Imagenet100, and ImageNet-R across different task granularities. The results show that sLoTh maintains strong performance even as the number of tasks increases.

\begin{table*}[htbp]
\centering
\caption{CIFAR100 CIL comparison between pretrained ViT with PEFT Continual Learning methods v/s pretrained QKFormer \& SpikingFormer on \textbf{Full sLoTh} for different task granularity, i.e., $T$ represents the number of tasks. Parameters are in millions. \textbf{First} and \underline{Second} best models are highlighted.}
\resizebox{\textwidth}{!}{
\begin{tabular}{lccccc}
\hline
\textbf{Methods} & \textbf{T=10} & \textbf{T=20} & \textbf{T=50} & \textbf{$\#$ Total Param.} & \textbf{$\#$updates} \\
\hline
L2P \cite{l2p}& 83.18$\pm$1.20 & 79.51$\pm$0.67 & 67.95$\pm$2.12 & 172 & \textbf{0.12} \\
DualPrompt \cite{wang2022dualprompt}& 81.48$\pm$0.86 & 80.44$\pm$1.38 & 72.5$\pm$1.08 & 172 & 0.86 \\
CODA-Prompt \cite{smith2023coda} & 86.31$\pm$0.12 & \underline{81.36$\pm$0.88} & \underline{73.77$\pm$0.98} & 172 & 4.6 \\
InfLoRA \cite{liang2024inflora}& \underline{86.75$\pm$0.35} & 80.97$\pm$0.74 & 70.68$\pm$1.26 & 172 & 0.51 \\
SD-LoRA \cite{sdlora}& \textbf{88.01$\pm$0.31} & OOM & OOM & 172 & \underline{0.39} \\
\textbf{Full sLoTh} & 81.23$\pm$1.74 & \textbf{84.65$\pm$1.46} & \textbf{87.22$\pm$1.59} & \textbf{64.32} & {0.44} \\
\textbf{Full sLoTh on SpikingFormer} &  63.28$\pm$1.39 & {67.23$\pm$ 2.03} & 70.48$\pm$1.21 & 66.34 & {0.56} \\
\hline
\end{tabular}}
\label{tab:cil_results_extended}
\end{table*}

\subsection{Online Continual Learning}

% We evaluate simplified variants of the proposed framework on the ImageNet-100 benchmark with 50 tasks. Table~\ref{tab:imagenet100_extended} reports the performance of sLoTh when only seLoRA or only threshold modulation is used. 
% % Interestingly, achieve higher accuracy, with \textbf{seLoRA-only} reaching \textbf{70.00$\pm$3.85} and \textbf{threshold-only} reaching \textbf{66.91$\pm$1.56}. This behaviour is consistent with our observations in the online continual learning (OCL) setting, where highly plastic mechanisms can adapt quickly in streaming scenarios.
% \begin{table}[htbp]
% \centering
% \caption{Extended comparison on ImageNet-100 with 50 tasks. Accuracy reported as mean $\pm$ std over 3 runs.}
% \begin{tabular}{l|c}
% \hline
% Method & Accuracy (\%) \\
% \hline
% L2P  & 56.25 $\pm$ 0.45 \\
% MVP & 54.33 $\pm$ 3.05 \\
% Online LoRA & 60.38 $\pm$ 2.59 \\
% \hline
% Full sLoTh & 62.14 $\pm$ 1.30 \\
% sLoTh (Threshold only) & 66.91 $\pm$ 1.56 \\
% sLoTh (seLoRA only) & \textbf{70.00 $\pm$ 3.85} \\
% \hline
% \end{tabular}
% \label{tab:imagenet100_extended}
% \end{table}
% Interestingly, the model achieves higher accuracy, with \textbf{seLoRA-only} reaching \textbf{70.00$\pm$3.85} and \textbf{threshold-only} reaching \textbf{66.91$\pm$1.56}. This behaviour is consistent with our observations in the online continual learning (OCL) setting given in the main paper's Table 4, where highly plastic mechanisms can adapt quickly in streaming scenarios.

We also evaluate the \textbf{seLoRA-only} variant of our method on a pretrained ViT-B/16 backbone (86M parameters) for 100 tasks on Tiny-ImageNet. The model achieves \textbf{55.68 $\pm$ 0.97} accuracy, indicating that the proposed adaptation mechanism generalises to non-spiking architectures.
% \section{Hardware and Software Environment}
% \label{appendix: hardware and software environment}

% All experiments were conducted on NVIDIA RTX A6000 and RTX A5000 GPUs utilising CUDA 12.8 for hardware acceleration. The training infrastructure consists of a workstation running Ubuntu 22.04.5 LTS with Python 3.13.11. Our implementation is based on PyTorch 2.10.0 and torchvision 0.25.0. To improve training stability and computational efficiency, we employed mixed precision training via automatic mixed precision (AMP).

% \section*{Acknowledgments}
% This was was supported in part by......

\bibliographystyle{unsrtnat}
\bibliography{references}

\end{document}